\documentclass[conference]{IEEEtran}
\IEEEoverridecommandlockouts
\usepackage{cite}
\usepackage{amsmath,amssymb,amsfonts}
\usepackage{algorithmic}
\usepackage{graphicx}
\usepackage{textcomp}
\usepackage{xcolor}
\usepackage{algorithm}
\usepackage{booktabs}
\usepackage{multirow}

\def\BibTeX{{\rm B\kern-.05em{\sc i\kern-.025em b}\kern-.08em
    T\kern-.1667em\lower.7ex\hbox{E}\kern-.125emX}}
\begin{document}

\title{FedUAF: Uncertainty-Aware Fusion with Reliability-Guided Aggregation for Multimodal Federated Sentiment Analysis}

\author{
Xianxun Zhu$^{1,4,6}$,
Zezhong Sun$^{2}$,
Imad Rida$^{3}$,
Erik Cambria$^{4}$,
Junqi Su$^{5}$,
Rui Wang$^{1}$,
Hui Chen$^{6,*}$\\

$^{1}$Shanghai University, China\\
$^{2}$North China Electric Power University, China\\
$^{3}$Université de Technologie de Compiègne, France\\
$^{4}$Nanyang Technological University, Singapore\\
$^{5}$City University of Hong Kong, Hong Kong SAR\\
$^{6}$Macquarie University, Australia
}

\maketitle

\begin{abstract}
Multimodal sentiment analysis in federated learning environments faces significant challenges due to missing modalities, heterogeneous data distributions, and unreliable client updates. Existing federated approaches often struggle to maintain robust performance under these practical conditions. In this paper, we propose FedUAF, a unified multimodal federated learning framework that addresses these challenges through uncertainty-aware fusion and reliability-guided aggregation. FedUAF explicitly models modality-level uncertainty during local training and leverages client reliability to guide global aggregation, enabling effective learning under incomplete and noisy multimodal data. Extensive experiments on CMU-MOSI and CMU-MOSEI demonstrate that FedUAF consistently outperforms state-of-the-art federated baselines across various missing-modality patterns and Non-IID settings. Moreover, FedUAF exhibits superior robustness against noisy clients, highlighting its potential for real-world multimodal federated applications.
\end{abstract}


\section{Introduction}

Multimodal affective analysis aims to infer human emotions and sentiments by jointly modeling heterogeneous signals such as visual expressions, vocal prosody, and linguistic content~\cite{liang2025modality,zhu2025client,wang2025raft}. As a core problem in multimedia intelligence, it underpins a wide range of real-world applications, including social media understanding, human-computer interaction, mental health monitoring, and intelligent recommendation systems. Despite substantial progress in centralized multimodal learning, practical deployments increasingly face stringent privacy constraints and distributed data ownership, motivating the adoption of collaborative and federated learning paradigms for affective computing.

Federated learning (FL) enables multiple clients to collaboratively train models without sharing raw data, making it particularly appealing for privacy-sensitive multimodal applications~\cite{chen2025fedsi,chen2025federated,nguyen2025learning}. However, applying FL to multimodal affective analysis remains highly challenging. First, affective data across clients are inherently non-IID, as emotional expressions vary significantly across individuals, cultures, recording environments, and interaction contexts~\cite{zhang2025generative}. Second, real-world multimedia systems frequently suffer from missing or unreliable modalities due to sensor failures, bandwidth limitations, or user-controlled data sharing, which severely degrades multimodal fusion performance~\cite{feng2023fedmultimodal}. Third, affective predictions are intrinsically subjective and uncertain, yet most existing multimodal federated learning methods treat modality contributions and client updates in a task-agnostic manner, ignoring emotion-related uncertainty that is crucial for robust sentiment inference~\cite{zhu2024emotion}.

Recent advances in multimodal federated learning have explored modality decoupling, split learning, and cross-modal aggregation to address data heterogeneity~\cite{chen2024feddat,wang2024fedmmr}. Nevertheless, most existing approaches either assume complete modality availability or rely on explicit modality reconstruction and generation, which introduce substantial computational overhead and are often ill-suited for resource-constrained edge devices. Moreover, current aggregation strategies typically weight client updates uniformly or based on data volume, without accounting for the reliability of affective predictions under heterogeneous emotional distributions and missing-modality conditions. These limitations hinder the deployment of scalable, trustworthy, and personalized distributed affective intelligence in real-world multimedia systems.

To address these challenges, we propose \textbf{FedUAF}, a \textbf{Federated Uncertainty-Aware Affective Fusion} framework for multimodal sentiment analysis under missing modalities and non-IID data distributions. FedUAF explicitly models emotion-related uncertainty at the modality level and leverages it to guide multimodal fusion, enabling the model to dynamically emphasize reliable modalities while suppressing noisy or missing ones without explicit modality reconstruction. Furthermore, FedUAF adopts a decomposed federated architecture that separates globally shared affective representations from client-specific decision heads, effectively balancing collaborative learning and personalized emotional expression. At the server side, FedUAF incorporates uncertainty-aware aggregation to weight client updates by affective reliability, improving robustness against noisy clients while maintaining communication efficiency.

We summarize our main contributions as follows:
\begin{itemize}
    \item We formulate federated multimodal affective learning under missing modalities, explicitly characterizing the joint challenges of modality heterogeneity, non-IID emotional distributions, and subjectivity-aware personalization in distributed multimedia systems.
    \item We propose FedUAF, a Federated Uncertainty-Aware Affective Fusion framework that leverages emotion-related uncertainty to dynamically weight multimodal representations, enabling robust sentiment inference without explicit modality reconstruction or generation.
    \item We design a decomposed federated architecture for affective personalization, separating globally shared affective representations from client-specific decision heads to balance collaborative learning and individualized emotional expression.
    \item We introduce an uncertainty-aware federated aggregation strategy, where client updates are weighted by affective reliability, improving robustness under missing-modality and noisy-client scenarios while preserving communication efficiency.
    \item Extensive experiments on standard multimodal sentiment benchmarks demonstrate that FedUAF consistently outperforms state-of-the-art multimodal federated learning baselines in accuracy, robustness to missing modalities, and scalability.
\end{itemize}

\section{Problem Formulation}

We consider a federated multimodal affective learning setting with $K$ clients collaboratively training a sentiment analysis model without sharing raw data. Each client corresponds to an edge device or data owner and possesses locally collected multimodal affective data with potentially missing modalities.

\subsection{Federated Multimodal Data}

Let $\mathcal{M}=\{v,a,t\}$ denote the set of modalities, corresponding to visual, audio, and textual inputs, respectively. The local dataset at client $k$ is denoted as
\begin{equation}
\mathcal{D}_k = \left\{ ( \mathbf{x}_{k,i}, y_{k,i} ) \right\}_{i=1}^{N_k},
\end{equation}
where $N_k$ is the number of samples at client $k$, $\mathbf{x}_{k,i} = \{ x_{k,i}^{m} \mid m \in \mathcal{M} \}$ represents the multimodal input features, and $y_{k,i}$ denotes the corresponding affective label (e.g., sentiment score or emotion category).

In real-world scenarios, not all modalities are always available. We define a binary modality availability mask
\begin{equation}
\mathbf{r}_{k,i} = \{ r_{k,i}^{m} \in \{0,1\} \mid m \in \mathcal{M} \},
\end{equation}
where $r_{k,i}^{m}=1$ indicates that modality $m$ is available for sample $i$ at client $k$, and $r_{k,i}^{m}=0$ denotes a missing modality. The observed multimodal input can thus be written as
\begin{equation}
\tilde{x}_{k,i}^{m} = r_{k,i}^{m} \cdot x_{k,i}^{m}.
\end{equation}

Due to user-specific behaviors, cultural differences, and heterogeneous data acquisition conditions, the local data distributions $\{\mathcal{D}_k\}_{k=1}^K$ are generally non-IID across clients.

\subsection{Prediction-Level Affective Uncertainty}

In multimodal affective analysis, prediction reliability varies across modalities and clients, especially under missing-modality and noisy data conditions. To capture this phenomenon, we introduce prediction-level affective uncertainty as a measure of the stability of affective predictions under stochastic perturbations.

Given an input sample $\mathbf{x}_{k,i}$, a local model produces a prediction $\hat{y}_{k,i}$. By performing multiple stochastic forward passes (e.g., with dropout enabled), we obtain a set of predictions $\{\hat{y}_{k,i}^{(1)}, \ldots, \hat{y}_{k,i}^{(T)}\}$. The affective uncertainty $u_{k,i}$ is computed as
\begin{equation}
u_{k,i} =
\begin{cases}
\mathrm{Var}\big(\hat{y}_{k,i}^{(1)}, \ldots, \hat{y}_{k,i}^{(T)}\big), & \text{regression}, \\
-\sum_c \bar{p}_{k,i}^{(c)} \log \bar{p}_{k,i}^{(c)}, & \text{classification},
\end{cases}
\end{equation}
where $\bar{p}_{k,i}$ denotes the averaged predictive probability over $T$ stochastic passes.

Importantly, the uncertainty defined above reflects prediction stability rather than parameter-level uncertainty and does not rely on Bayesian inference.

\subsection{Federated Learning Objective}

The goal of federated multimodal affective learning is to optimize a global model that generalizes well across heterogeneous clients while respecting data privacy. Let $\theta_s$ denote the parameters of the globally shared representation and $\theta_k$ the client-specific parameters at client $k$. The federated optimization objective can be formulated as
\begin{equation}
\min_{\theta_s, \{\theta_k\}} \sum_{k=1}^{K} \mathbb{E}_{(\mathbf{x},y)\sim \mathcal{D}_k}
\left[ \mathcal{L}\big(f(\mathbf{x}; \theta_s, \theta_k), y\big) \right],
\end{equation}
where $\mathcal{L}(\cdot)$ denotes the affective prediction loss.

In this work, affective uncertainty is leveraged as a \emph{reliability signal} to guide multimodal fusion and federated aggregation, enabling robust collaborative learning under missing modalities and non-IID client distributions. The detailed framework and optimization procedure are introduced in the following section.

\section{Methodology}

\begin{figure}[t]
    \centering
    \includegraphics[width=\linewidth]{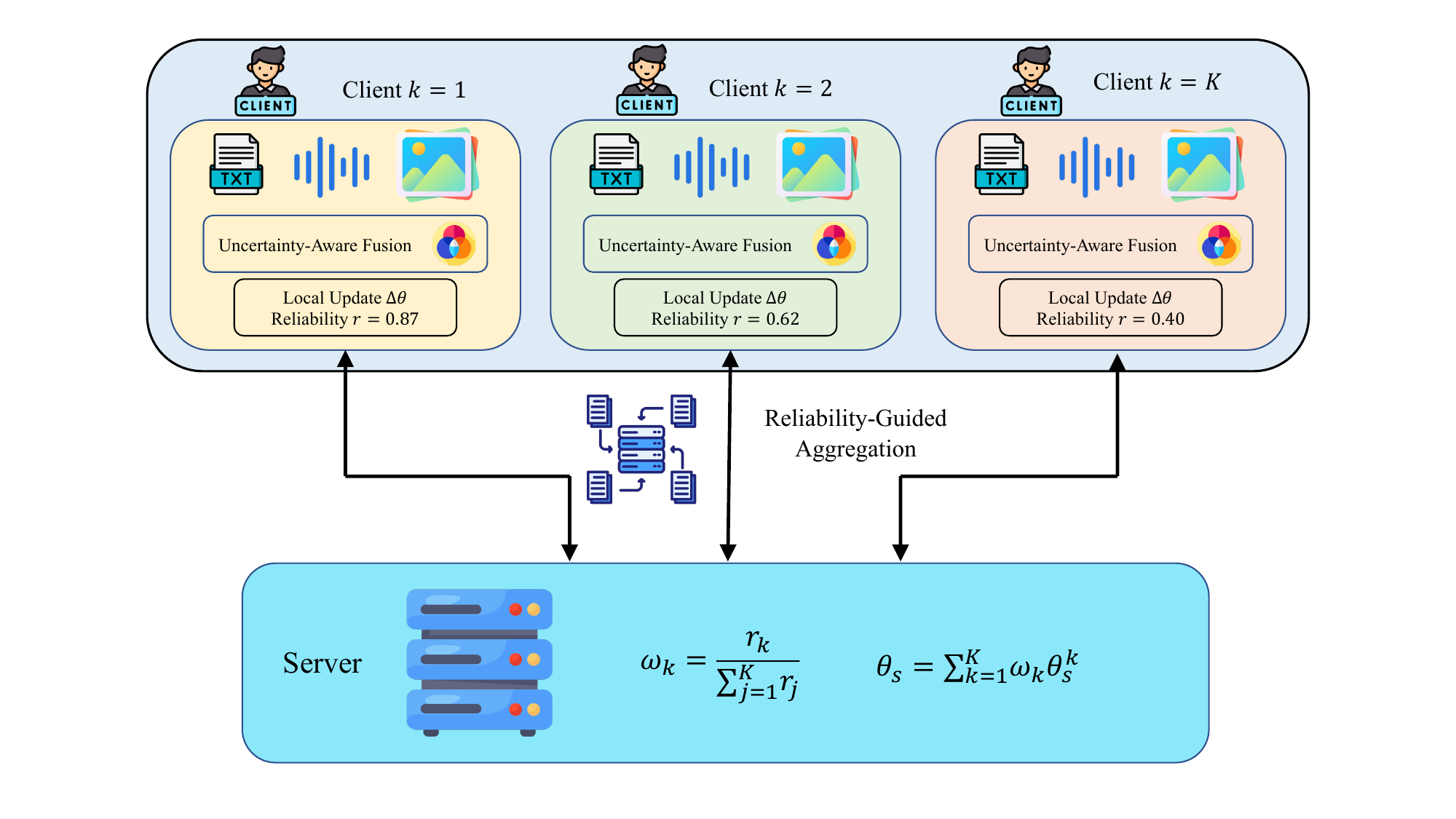}
    \caption{Overview of the proposed FedUAF framework.}
    \label{fig:feduaf_framework}
\end{figure}

In this section, we present the proposed FedUAF framework, a Federated Uncertainty-Aware Affective Fusion approach for robust multimodal sentiment analysis under missing modalities and heterogeneous client distributions. FedUAF explicitly leverages prediction-level affective uncertainty to guide multimodal fusion at the client side and federated aggregation at the server side, while maintaining a deterministic learning paradigm. Fig.~\ref{fig:feduaf_framework} presents an overview of the proposed FedUAF framework.

\subsection{Model Architecture Overview}

FedUAF adopts a decomposed federated architecture consisting of a globally shared representation module and client-specific decision heads. Specifically, each client maintains modality-specific encoders
\begin{equation}
h_{k,i}^{m} = E^{m}(x_{k,i}^{m}), \quad m \in \mathcal{M},
\end{equation}
where $E^{m}(\cdot)$ denotes the encoder for modality $m$. The encoders share the same architecture across clients but do not require parameter sharing, allowing flexibility under heterogeneous data conditions.

The extracted modality representations are then fused into a unified affective representation, which is passed through a shared affective representation head and a client-specific prediction head for sentiment inference.

\subsection{Prediction-Level Uncertainty Estimation}

To assess the reliability of affective predictions, FedUAF estimates uncertainty at the prediction level using stochastic forward passes. For each input sample, we perform $T$ stochastic forward passes with dropout enabled, yielding a set of predictions $\{\hat{y}_{k,i}^{(1)}, \ldots, \hat{y}_{k,i}^{(T)}\}$.

The affective uncertainty is computed as
\begin{equation}
u_{k,i} =
\begin{cases}
\mathrm{Var}\big(\hat{y}_{k,i}^{(1)}, \ldots, \hat{y}_{k,i}^{(T)}\big), & \text{regression}, \\
-\sum_c \bar{p}_{k,i}^{(c)} \log \bar{p}_{k,i}^{(c)}, & \text{classification},
\end{cases}
\end{equation}
where $\bar{p}_{k,i}$ denotes the averaged predictive probability over $T$ stochastic passes.

The estimated uncertainty reflects the stability of affective predictions under stochastic perturbations and serves as a lightweight reliability signal. Notably, FedUAF does not perform Bayesian inference or model parameter posteriors.

\subsection{Uncertainty-Guided Affective Fusion}

Given modality-specific representations $\{h_{k,i}^{m}\}$ and their corresponding uncertainties $\{u_{k,i}^{m}\}$, FedUAF performs uncertainty-guided affective fusion by assigning higher weights to more reliable modalities.

We compute the modality fusion weight as
\begin{equation}
\alpha_{k,i}^{m} = \frac{\exp(-u_{k,i}^{m})}{\sum_{m' \in \mathcal{M}} \exp(-u_{k,i}^{m'})},
\end{equation}
where unavailable modalities are masked by setting their weights to zero.

The fused affective representation is obtained as
\begin{equation}
h_{k,i} = \sum_{m \in \mathcal{M}} \alpha_{k,i}^{m} \cdot h_{k,i}^{m}.
\end{equation}

This mechanism enables the model to dynamically emphasize reliable modalities while suppressing noisy or missing ones, without requiring explicit modality reconstruction or generation.

\subsection{Uncertainty-Aware Federated Aggregation}

To enhance robustness against noisy clients and heterogeneous data distributions, FedUAF incorporates uncertainty-aware federated aggregation. Each client computes a client-level affective uncertainty score by averaging prediction-level uncertainties over its local dataset:
\begin{equation}
\bar{u}_k = \frac{1}{N_k} \sum_{i=1}^{N_k} u_{k,i}.
\end{equation}

Affective reliability is then defined as
\begin{equation}
r_k = \frac{1}{\bar{u}_k + \epsilon},
\end{equation}
where $\epsilon$ is a small constant for numerical stability.

During federated aggregation, the server aggregates the shared representation parameters $\{\theta_s^k\}$ using reliability-weighted averaging:
\begin{equation}
\theta_s = \sum_{k=1}^{K} w_k \theta_s^k, \quad
w_k = \frac{r_k}{\sum_{j=1}^{K} r_j}.
\end{equation}

Importantly, FedUAF aggregates only the shared model parameters and does not aggregate uncertainty statistics as model parameters. The uncertainty estimates are used solely to guide the weighting of client updates, resulting in communication complexity comparable to standard FedAvg.

\subsection{Training Procedure}

FedUAF follows a standard synchronous federated learning protocol. In each communication round, the server broadcasts the shared parameters to selected clients. Each client performs local training using uncertainty-guided fusion and uploads the updated shared parameters along with a scalar reliability score. The server then performs uncertainty-aware aggregation to update the global model. The overall training procedure is summarized in Algorithm~\ref{alg:feduaf}.

\begin{algorithm}[t]
\caption{Federated Training of FedUAF}
\label{alg:feduaf}
\begin{algorithmic}[1]
\REQUIRE Number of communication rounds $R$, client set $\mathcal{K}$, local epochs $E$, learning rate $\eta$
\ENSURE Optimized shared parameters $\theta_s$
\STATE Initialize shared parameters $\theta_s$
\FOR{$r = 1, \dots, R$}
    \STATE \textbf{Server:} Broadcast shared parameters $\theta_s$ to selected clients
    \FOR{\textbf{each client} $k \in \mathcal{K}$ \textbf{in parallel}}
        \STATE Initialize local shared parameters $\theta_s^k \leftarrow \theta_s$
        \FOR{$e = 1, \dots, E$}
            \FOR{\textbf{each local minibatch} $(\mathbf{x}, y)$}
                \STATE Extract modality-specific representations $\{h^{m}\}_{m \in \mathcal{M}}$
                \STATE Estimate prediction-level uncertainty $\{u^{m}\}_{m \in \mathcal{M}}$
                \STATE Compute uncertainty-guided fusion weights $\{\alpha^{m}\}_{m \in \mathcal{M}}$
                \STATE Fuse multimodal representations $h \leftarrow \sum_{m} \alpha^{m} h^{m}$
                \STATE Compute affective prediction $\hat{y} \leftarrow f(h)$
                \STATE Update local parameters using gradient descent with step size $\eta$
            \ENDFOR
        \ENDFOR
        \STATE Estimate client-level uncertainty $\bar{u}_k$ over local data
        \STATE Compute affective reliability score $r_k \leftarrow (\bar{u}_k + \epsilon)^{-1}$
        \STATE Upload updated shared parameters $\theta_s^k$ and reliability $r_k$ to server
    \ENDFOR
    \STATE \textbf{Server:} Normalize reliability scores $\{r_k\}$
    \STATE \textbf{Server:} Aggregate shared parameters using reliability-weighted averaging
\ENDFOR
\end{algorithmic}
\end{algorithm}

\section{Experiments}

In this section, we evaluate the effectiveness of the proposed FedUAF framework on multimodal affective analysis tasks. We aim to answer the following questions:  
(1) Does FedUAF outperform existing multimodal federated learning baselines under missing-modality settings?  
(2) How robust is FedUAF to non-IID client distributions and noisy clients?  
(3) What is the contribution of uncertainty-aware fusion and aggregation?

\subsection{Datasets}

We conduct experiments on two widely used multimodal sentiment analysis benchmarks:

\textbf{CMU-MOSI}~\cite{huang2023dominant} is a multimodal opinion sentiment dataset consisting of short video clips annotated with sentiment scores in the range $[-3,3]$. Each sample contains synchronized visual, audio, and textual modalities.

\textbf{CMU-MOSEI}~\cite{zadeh2018multimodal} is a large-scale extension of MOSI, covering a broader range of speakers and topics. It includes over 23,000 annotated video segments with multimodal inputs and sentiment labels.

Following common practice, we use the standard train/validation/test splits provided by the datasets. Sentiment prediction is formulated as a regression task.

\subsection{Federated Data Partition}

To simulate federated learning scenarios, we partition the datasets by speaker identity, where each speaker corresponds to one client. This partition strategy naturally induces non-IID data distributions across clients due to speaker-specific emotional expression styles.

Missing-modality scenarios are simulated by randomly dropping one or more modalities at the sample level with a predefined probability. Once a modality is missing, it is treated as unavailable during both training and inference.

\subsection{Baselines}

We compare FedUAF with the following representative baselines:

\textbf{FedAvg}~\cite{mcmahan2017communication}: The standard federated averaging algorithm with simple multimodal feature concatenation.

\textbf{FedProx}~\cite{li2020federated}: A federated optimization method designed to handle heterogeneous client distributions.

\textbf{FedMSplit}~\cite{chen2022fedmsplit}: A multimodal federated learning approach that decouples modality-specific encoders and shared representations.

\textbf{MIFL}~\cite{phung2024contrastive}: A multimodal federated learning framework that models cross-modal interactions under data heterogeneity.

\textbf{FedMAC}~\cite{nguyen2024fedmac}: A state-of-the-art multimodal federated learning method that addresses modality heterogeneity via adaptive aggregation.

All baselines are implemented under the same federated setting and trained using identical data partitions for fair comparison.

\subsection{Evaluation Metrics}

For sentiment regression, we adopt Mean Absolute Error (MAE) as the evaluation metric, which is widely used in multimodal affective analysis. MAE directly measures the average absolute difference between predicted sentiment scores and ground-truth labels, providing an intuitive and robust assessment of regression performance. All results are reported as the average MAE over all clients in the federated test set.

\subsection{Implementation Details}

All models are implemented in PyTorch. Visual features are extracted using a pre-trained CNN backbone, audio features using a convolutional audio encoder, and textual features using a lightweight Transformer-based encoder. The hidden dimension of modality representations is set to 128.

We train the models for $R=100$ communication rounds, with $E=5$ local epochs per round. The Adam optimizer is used with an initial learning rate of $1\times10^{-3}$. For uncertainty estimation, we perform $T=5$ stochastic forward passes with dropout enabled. All experiments are conducted on a workstation equipped with an NVIDIA Tesla V100 GPU (60GB).

\subsection{Main Results}

We evaluate the proposed FedUAF under varying degrees of data heterogeneity and modality incompleteness to reflect realistic federated affective computing scenarios. Data heterogeneity is controlled by the Non-IID intensity ranging from 0.2 to 1.0, while modality incompleteness is simulated using missing-modality ratios $\rho_m \in \{0.2, 0.8\}$, corresponding to mild and severe modality absence.

The quantitative results on CMU-MOSI and CMU-MOSEI are summarized in Table~\ref{tab:main_results_mae}. As expected, increasing either the Non-IID intensity or the missing-modality ratio consistently degrades the performance of all methods. Nevertheless, FedUAF achieves the lowest MAE across all evaluated settings, with more pronounced advantages under higher heterogeneity and severe modality absence. Compared with representative federated multimodal baselines, FedUAF demonstrates improved robustness and stability, validating the effectiveness of uncertainty-aware fusion and reliability-guided aggregation in challenging federated multimodal environments.

\begin{table*}[t]
\centering
\caption{Performance (MAE $\downarrow$) under different missing-modality ratios in both IID and Non-IID settings. Results are reported as mean $\pm$ standard deviation over 3 runs. Best and second-best results are highlighted in \textbf{bold} and \underline{underline}, respectively.}
\label{tab:main_results_mae}
\resizebox{\textwidth}{!}{
\begin{tabular}{l c l | ccccc | ccccc}
\toprule
\multirow{2}{*}{Dataset} & \multirow{2}{*}{$\rho_m$} & \multirow{2}{*}{Method}
& \multicolumn{5}{c|}{IID} 
& \multicolumn{5}{c}{Non-IID} \\
& & & 0.2 & 0.4 & 0.6 & 0.8 & 1.0 & 0.2 & 0.4 & 0.6 & 0.8 & 1.0 \\
\midrule

\multirow{12}{*}{MOSI}
& \multirow{6}{*}{0.2}
& FedAvg    & 0.88$\pm$0.06 & 0.90$\pm$0.07 & 0.91$\pm$0.07 & 0.96$\pm$0.04 & 0.96$\pm$0.06
          & 0.90$\pm$0.08 & 0.95$\pm$0.07 & 1.01$\pm$0.02 & 1.07$\pm$0.05 & 1.13$\pm$0.09 \\
& & FedProx   & \underline{0.82$\pm$0.04} & 0.89$\pm$0.04 & 0.90$\pm$0.13 & 0.95$\pm$0.05 & 0.98$\pm$0.04
          & 0.88$\pm$0.06 & 0.93$\pm$0.09 & 0.98$\pm$0.04 & 1.04$\pm$0.11 & 1.10$\pm$0.07 \\
& & FedMSplit & 0.85$\pm$0.04 & 0.88$\pm$0.03 & 0.91$\pm$0.09 & \textbf{0.89$\pm$0.04} & 0.97$\pm$0.04
          & 0.89$\pm$0.02 & 0.91$\pm$0.03 & 0.96$\pm$0.03 & 1.02$\pm$0.04 & 1.07$\pm$0.08 \\
& & MIFL      & 0.84$\pm$0.02 & 0.87$\pm$0.03 & 0.90$\pm$0.03 & 0.93$\pm$0.04 & 0.96$\pm$0.09
          & 0.85$\pm$0.11 & 0.90$\pm$0.03 & 0.99$\pm$0.06 & 1.06$\pm$0.04 & 1.05$\pm$0.05 \\
& & FedMAC    & 0.83$\pm$0.02 & \underline{0.86$\pm$0.03} & \underline{0.89$\pm$0.06} & 0.92$\pm$0.03 & \underline{0.95$\pm$0.04}
          & \underline{0.84$\pm$0.02} & \underline{0.88$\pm$0.11} & \underline{0.93$\pm$0.03} & \underline{0.98$\pm$0.04} & \underline{1.02$\pm$0.04} \\
& & \textbf{FedUAF} & \textbf{0.79$\pm$0.02} & \textbf{0.82$\pm$0.02} & \textbf{0.85$\pm$0.03} & \underline{0.90$\pm$0.03} & \textbf{0.91$\pm$0.03}
          & \textbf{0.80$\pm$0.02} & \textbf{0.84$\pm$0.02} & \textbf{0.88$\pm$0.03} & \textbf{0.92$\pm$0.03} & \textbf{0.96$\pm$0.04} \\
\cmidrule(lr){2-13}
& \multirow{6}{*}{0.8}
& FedAvg    & \underline{0.94$\pm$0.03} & 1.02$\pm$0.13 & 1.06$\pm$0.09 & 1.10$\pm$0.05 & 1.09$\pm$0.04
          & 1.02$\pm$0.09 & 1.11$\pm$0.04 & 1.18$\pm$0.05 & 1.25$\pm$0.03 & 1.32$\pm$0.03 \\
& & FedProx   & 0.97$\pm$0.04 & 0.97$\pm$0.08 & 1.04$\pm$0.12 & 1.08$\pm$0.06 & 1.12$\pm$0.03
          & 1.03$\pm$0.07 & \underline{1.02$\pm$0.05} & 1.15$\pm$0.03 & 1.22$\pm$0.02 & 1.28$\pm$0.07 \\
& & FedMSplit & 0.96$\pm$0.11 & 0.99$\pm$0.06 & 1.07$\pm$0.07 & 1.04$\pm$0.11 & 1.11$\pm$0.06
          & 1.01$\pm$0.11 & 1.06$\pm$0.11 & 1.16$\pm$0.02 & 1.18$\pm$0.06 & 1.24$\pm$0.06 \\
& & MIFL      & 0.95$\pm$0.06 & 1.03$\pm$0.04 & 1.05$\pm$0.07 & 1.06$\pm$0.05 & 1.09$\pm$0.06
          & 0.99$\pm$0.08 & 1.05$\pm$0.12 & 1.10$\pm$0.08 & 1.16$\pm$0.11 & 1.22$\pm$0.06 \\
& & FedMAC    & \textbf{0.93$\pm$0.04} & \underline{0.96$\pm$0.04} & \underline{1.00$\pm$0.04} & \underline{1.03$\pm$0.02} & \underline{1.07$\pm$0.05}
          & \underline{0.97$\pm$0.03} & 1.03$\pm$0.04 & \underline{1.07$\pm$0.05} & \underline{1.12$\pm$0.05} & \underline{1.17$\pm$0.06} \\
& & \textbf{FedUAF} & 0.95$\pm$0.09 & \textbf{0.93$\pm$0.03} & \textbf{0.96$\pm$0.04} & \textbf{1.00$\pm$0.04} & \textbf{1.03$\pm$0.05}
          & \textbf{0.90$\pm$0.03} & \textbf{0.95$\pm$0.03} & \textbf{1.00$\pm$0.04} & \textbf{1.05$\pm$0.05} & \textbf{1.10$\pm$0.05} \\

\midrule

\multirow{12}{*}{MOSEI}
& \multirow{6}{*}{0.2}
& FedAvg    & 0.70$\pm$0.05 & 0.67$\pm$0.13 & 0.72$\pm$0.02 & 0.76$\pm$0.07 & 0.76$\pm$0.08
          & 0.78$\pm$0.06 & 0.82$\pm$0.03 & 0.86$\pm$0.04 & 0.91$\pm$0.05 & 0.96$\pm$0.06 \\
& & FedProx   & \underline{0.66$\pm$0.07} & 0.71$\pm$0.06 & 0.75$\pm$0.09 & 0.73$\pm$0.14 & 0.82$\pm$0.03
          & 0.75$\pm$0.04 & 0.80$\pm$0.08 & 0.84$\pm$0.02 & 0.85$\pm$0.05 & 0.89$\pm$0.03 \\
& & FedMSplit & 0.68$\pm$0.04 & 0.70$\pm$0.08 & 0.72$\pm$0.03 & 0.74$\pm$0.11 & 0.76$\pm$0.06
          & 0.76$\pm$0.07 & 0.81$\pm$0.13 & 0.85$\pm$0.06 & \textbf{0.79$\pm$0.09} & 0.90$\pm$0.05 \\
& & MIFL      & 0.67$\pm$0.02 & \textbf{0.64$\pm$0.05} & 0.71$\pm$0.09 & 0.73$\pm$0.03 & 0.80$\pm$0.04
          & 0.73$\pm$0.02 & 0.77$\pm$0.11 & 0.81$\pm$0.04 & 0.85$\pm$0.05 & 0.89$\pm$0.05 \\
& & FedMAC    & 0.69$\pm$0.02 & 0.68$\pm$0.02 & 0.70$\pm$0.03 & \underline{0.72$\pm$0.03} & \underline{0.74$\pm$0.03}
          & \underline{0.72$\pm$0.02} & \underline{0.75$\pm$0.07} & \underline{0.79$\pm$0.03} & \underline{0.83$\pm$0.04} & \underline{0.87$\pm$0.11} \\
& & \textbf{FedUAF} & \textbf{0.63$\pm$0.04} & \underline{0.65$\pm$0.02} & \textbf{0.67$\pm$0.03} & \textbf{0.69$\pm$0.03} & \textbf{0.71$\pm$0.03}
          & \textbf{0.68$\pm$0.07} & \textbf{0.71$\pm$0.04} & \textbf{0.75$\pm$0.03} & 0.85$\pm$0.03 & \textbf{0.83$\pm$0.04} \\
\cmidrule(lr){2-13}
& \multirow{6}{*}{0.8}
& FedAvg    & 0.73$\pm$0.02 & 0.80$\pm$0.03 & 0.82$\pm$0.03 & 0.84$\pm$0.06 & 0.83$\pm$0.07
          & 0.83$\pm$0.06 & 0.90$\pm$0.04 & 0.94$\pm$0.03 & 0.96$\pm$0.07 & 1.04$\pm$0.09 \\
& & FedProx   & 0.75$\pm$0.04 & 0.82$\pm$0.08 & 0.81$\pm$0.11 & 0.83$\pm$0.01 & 0.83$\pm$0.07
          & 0.85$\pm$0.09 & 0.88$\pm$0.04 & 0.91$\pm$0.07 & 0.97$\pm$0.03 & 1.02$\pm$0.02 \\
& & FedMSplit & 0.76$\pm$0.02 & 0.78$\pm$0.02 & 0.80$\pm$0.01 & 0.85$\pm$0.02 & 0.84$\pm$0.09
          & 0.82$\pm$0.03 & 0.86$\pm$0.05& 0.95$\pm$0.05 & 0.95$\pm$0.09 & 1.02$\pm$0.12 \\
& & MIFL      & 0.75$\pm$0.06 & 0.77$\pm$0.09 & 0.79$\pm$0.07 & 0.81$\pm$0.04 & 0.83$\pm$0.04
          & 0.81$\pm$0.02 & 0.88$\pm$0.11 & 0.89$\pm$0.11 & 0.94$\pm$0.03 & 1.06$\pm$0.08 \\
& & FedMAC    & \underline{0.74$\pm$0.12} & \underline{0.76$\pm$0.03} & \underline{0.78$\pm$0.03} & \underline{0.80$\pm$0.03} & \underline{0.82$\pm$0.06}
          & \underline{0.79$\pm$0.02} & \underline{0.83$\pm$0.04} & \underline{0.87$\pm$0.05} & \textbf{0.89$\pm$0.05} & \underline{0.97$\pm$0.04} \\
& & \textbf{FedUAF} & \textbf{0.71$\pm$0.03} & \textbf{0.73$\pm$0.03} & \textbf{0.75$\pm$0.03} & \textbf{0.77$\pm$0.04} & \textbf{0.79$\pm$0.04}
          & \textbf{0.76$\pm$0.03} & \textbf{0.80$\pm$0.04} & \textbf{0.84$\pm$0.05} & \underline{0.92$\pm$0.05} & \textbf{0.94$\pm$0.04} \\
\bottomrule
\end{tabular}
}
\end{table*}

\subsection{Ablation Study}

\begin{figure}[!t]
    \centering
    \includegraphics[width=0.85\linewidth]{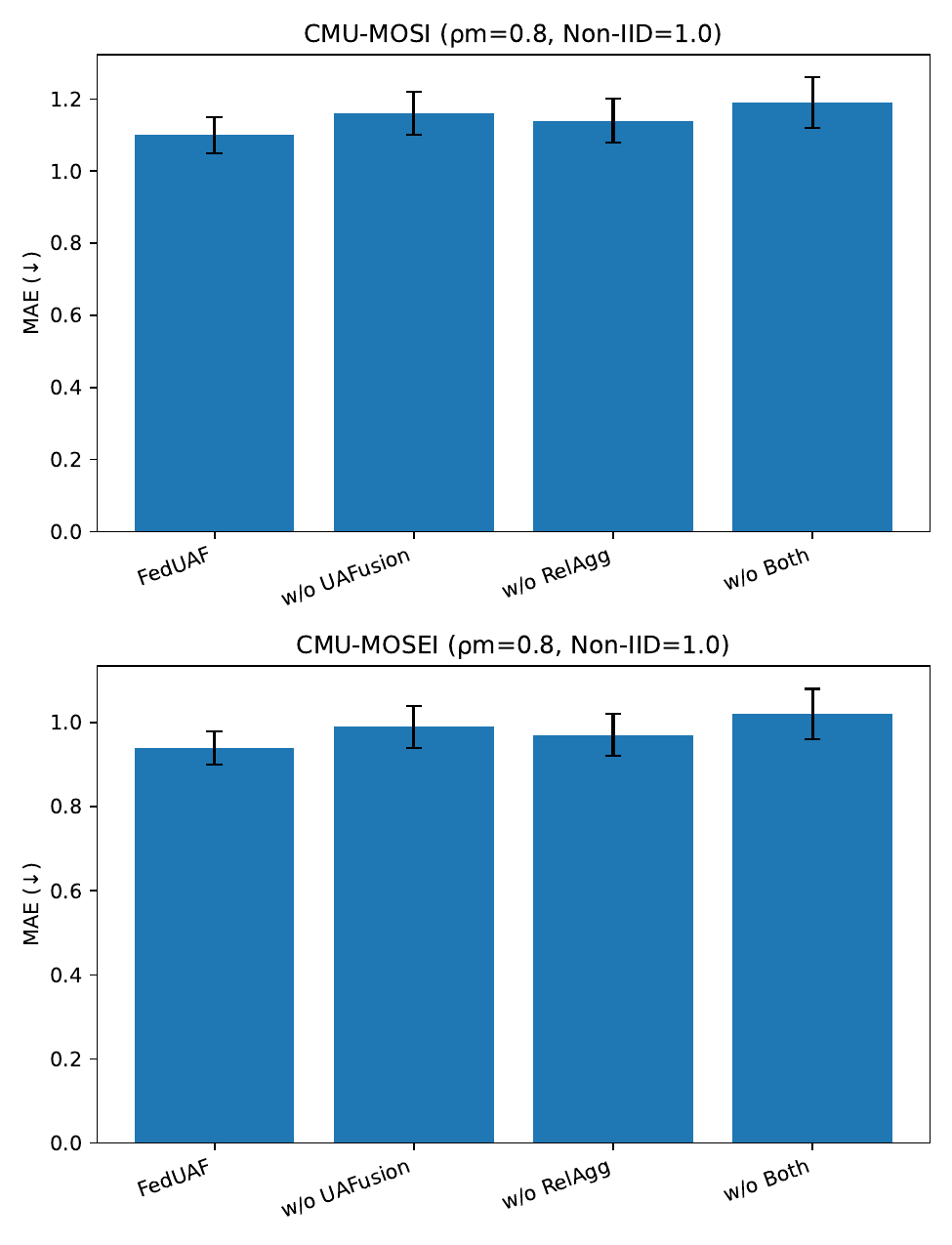}
    \caption{Ablation study of FedUAF under severe missing-modality ($\rho_m=0.8$) and strong data heterogeneity (Non-IID intensity $=1.0$). Results on CMU-MOSI (top) and CMU-MOSEI (bottom) are reported in terms of MAE (mean$\pm$std over 3 runs).}
    \label{fig:ablation}
\end{figure}

We conduct an ablation study to verify the contributions of the two key components in FedUAF: (i) uncertainty-aware fusion (UAFusion) and (ii) reliability-guided aggregation (RelAgg). We evaluate four variants under a challenging setting ($\rho_m=0.8$, Non-IID intensity $=1.0$): the full FedUAF, FedUAF w/o UAFusion, FedUAF w/o RelAgg, and FedUAF w/o both. As shown in Fig.~\ref{fig:ablation}, removing either component consistently degrades MAE on both CMU-MOSI and CMU-MOSEI, while removing both leads to the largest performance drop. These results confirm that UAFusion and RelAgg are complementary and jointly contribute to the robustness of FedUAF under severe modality incompleteness and strong client heterogeneity.

\subsection{Robustness to Noisy Clients}

In practical federated learning systems, client updates are often unreliable due to noisy labels, corrupted local data, or unstable training behaviors. To evaluate the robustness of FedUAF under such adverse conditions, we introduce noisy clients whose local updates are intentionally perturbed and vary the noisy client ratio from 0 to 0.6. Throughout this experiment, we fix the missing-modality ratio to $\rho_m=0.8$ and the Non-IID intensity to 1.0, representing a challenging federated setting with severe modality incompleteness and strong data heterogeneity.

\begin{figure}[!t]
    \centering
    \includegraphics[width=0.85\linewidth]{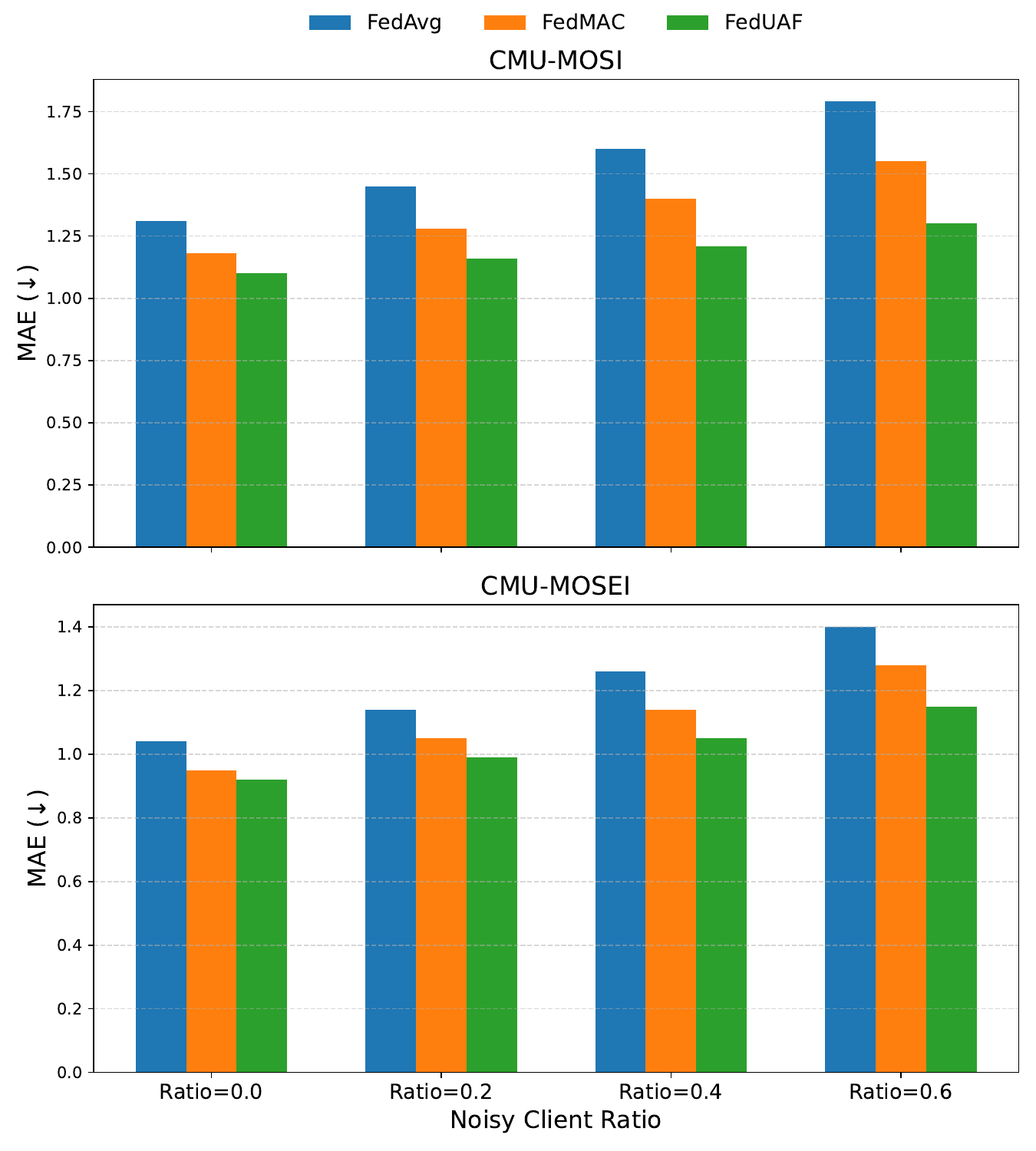}
    \caption{Robustness to noisy clients under severe missing-modality ($\rho_m=0.8$) and strong data heterogeneity (Non-IID intensity $=1.0$). Results on CMU-MOSI (top) and CMU-MOSEI (bottom) are reported in terms of MAE.}
    \label{fig:noisy_clients}
\end{figure}

As shown in Fig.~\ref{fig:noisy_clients}, the MAE of all methods increases as the proportion of noisy clients grows on both CMU-MOSI and CMU-MOSEI. However, FedUAF consistently achieves lower MAE and exhibits a noticeably slower performance degradation compared with FedAvg and FedMAC. In particular, the performance gap becomes more pronounced when the noisy client ratio exceeds 0.4, indicating that reliability-guided aggregation in FedUAF effectively suppresses the influence of unreliable client updates. These results demonstrate the robustness of FedUAF against noisy and potentially adversarial clients in challenging multimodal federated learning scenarios.

\section{Conclusion}

In this paper, we proposed FedUAF, a unified federated learning framework for multimodal sentiment analysis under missing-modality and heterogeneous data distributions. By incorporating uncertainty-aware fusion and reliability-guided aggregation, FedUAF effectively mitigates the negative impact of incomplete modalities, strong client heterogeneity, and noisy client updates. Extensive experiments on CMU-MOSI and CMU-MOSEI demonstrate that FedUAF consistently outperforms existing federated baselines across various missing-modality patterns and Non-IID settings, while exhibiting superior robustness under adverse federated conditions. These results highlight the potential of uncertainty- and reliability-aware design for building robust multimodal federated learning systems in real-world applications.

\bibliographystyle{IEEEbib}
\bibliography{icme2026references}

\end{document}